\documentclass[10pt,twocolumn,letterpaper]{article}

\usepackage{cvpr}
\usepackage{times}
\usepackage{epsfig}
\usepackage{graphicx}
\usepackage{amsmath}
\usepackage{amssymb}
\usepackage{xfrac}
\usepackage{ifthen}

\usepackage[pagebackref=true,breaklinks=true,letterpaper=true,colorlinks,bookmarks=false]{hyperref}

\DeclareSymbolFont{bbold}{U}{bbold}{m}{n}
\DeclareSymbolFontAlphabet{\mathbbold}{bbold}

\cvprfinalcopy %

\def\cvprPaperID{3425} %

\ifcvprfinal\fi
\begin{document}

\newcommand{\onehalf}{\frac{1}{2}}
\newcommand{\stimes}{\!\!\times\!\!}
\newcommand{\lessspace}[1]{#1\hspace{-0.2em}}
\newcommand{\onelessspace}{\lessspace{1}}
\newcommand{\id}[1]{\mbox{\footnotesize \emph{#1}}}
\newcommand{\littlex}{{\scriptsize \textsf{x}}}
\newcommand{\tred}[1]{\textcolor{red}{#1}}
\newcommand{\tbred}[1]{\textcolor{red}{\textbf{#1}}}
\newcommand{\Fhalf}{F\hspace{-0.2em}/\hspace{-0.1em}2}

\title{On the Importance of Stereo for Accurate Depth Estimation: \\
An Efficient Semi-Supervised Deep Neural Network Approach}

\author{Nikolai Smolyanskiy \qquad Alexey Kamenev \qquad Stan Birchfield \\
NVIDIA \\
{\tt\small \{nsmolyanskiy, akamenev, sbirchfield\}@nvidia.com}}%

\maketitle
\thispagestyle{empty}

\begin{abstract}
	We revisit the problem of visual depth estimation in the context of autonomous vehicles.
	Despite the progress on monocular depth estimation in recent years, we show
	that the gap between monocular and stereo depth accuracy remains large---a particularly relevant result due to the prevalent reliance upon monocular 
	cameras by vehicles that are expected to be self-driving.
	We argue that the challenges of removing this gap are significant, owing to fundamental limitations
	of monocular vision.  
	As a result, we focus our efforts on depth estimation by stereo.
	We propose a novel semi-supervised learning approach to training a deep stereo neural network, along with a novel architecture containing a machine-learned argmax layer and a custom runtime that enables a smaller version of our stereo DNN to run on an embedded GPU.  Competitive results are shown on the KITTI 2015 stereo dataset.  We also evaluate the recent progress of stereo algorithms
	by measuring the impact upon accuracy of various design criteria.\footnote{Video of the system is at \url{https://youtu.be/0FPQdVOYoAU}.} 
\end{abstract}

\section{Introduction}

Estimating depth from images is a long-standing problem in computer vision.  Depth perception is useful for scene understanding, scene reconstruction, virtual and augmented reality, obstacle avoidance, self-driving cars, robotics, and other applications.

Traditionally, multiple images have been used to estimate depth.  Techniques that fall within this category include stereo, photometric stereo, depth from focus, depth from defocus, time-of-flight,\footnote{Although time-of-flight does not, in theory require multiple images, in practice multiple images are collected with different bandwidths in order to achieve high accuracy over long ranges.} and structure from motion.  The reasons for using multiple images are twofold:  1) absolute depth estimates require at least one known distance in the world, which can often be provided by some knowledge regarding the multi-camera rig (e.g., the baseline between stereo cameras); and 2) multiple images provide geometric constraints that can be leveraged to overcome the many ambiguities of photometric data.

The alternative is to use a single image to estimate depth.  We argue that this alternative---due to its fundamental limitations---is not likely to be able to achieve high-accuracy depth estimation at large distances in unfamiliar environments.  As a result, in the context of self-driving cars we believe monocular depth estimation is not likely to yield results with sufficient accuracy.  In contrast, we offer a novel, efficient deep-learning stereo approach that achieves compelling results on the KITTI 2015 dataset by leveraging a semi-supervised loss function (using LIDAR and photometric consistency), concatenating cost volume, 3D convolutions, and a machine-learned argmax function.  The contributions of the paper are as follows:
\begin{itemize}
\setlength\itemsep{-0.25em}
	\item Quantitative and qualitative demonstration of the gap in depth accuracy between monocular and stereoscopic depth.
	\item A novel semi-supervised approach (combining lidar and photometric losses) to training a deep stereo neural network.  To our knowledge, ours is the first deep stereo network to do so.\footnote{Similarly, Kuznietsov et al. \cite{kuznietsov2017cvpr:mono} use a semi-supervised approach for training a monocular network.}
	\item A smaller version of our network, and a custom runtime, that runs at near real-time ($\sim$20 fps) on a standard GPU, and runs efficiently on an embedded GPU.  To our knowledge, ours is the first stereo DNN to run on an embedded GPU.
	\item Quantitative analysis of various network design choices, along with a novel machine-learned argmax layer that yields smoother disparity maps.
\end{itemize}

\section{Motivation}

The undeniable success of deep neural networks in computer vision has encouraged researchers to pursue the problem of estimating depth from a single image \cite{eigen2014nips:mono,liu2016pami:depth,garg2016eccv:cnn,godard2017cvpr:unsup,kuznietsov2017cvpr:mono}.  This is, no doubt, a noble endeavor:  if it were possible to accurately estimate depth from a single image, then the complexity (and hence cost) of the hardware needed would be dramatically reduced, which would broaden the applicability substantially.  An excellent overview of existing work on monocular depth estimation can be found in~\cite{godard2017cvpr:unsup}.

Nevertheless, there are reasons to be cautious about the reported success of monocular depth.  To date, monocular depth solutions, while yielding encouraging preliminary results, are not at the point where reliable information (from a robotics point of view) can be expected from them.  And although such solutions will continue to improve, monocular depth will never overcome well-known fundamental limitations, such as the need for a world measurement to infer absolute depth, and the ambiguity that arises when a photograph is taken of a photograph (an important observation for biometric and security systems).  

One of the motivations for monocular depth is a long-standing belief that stereo is only useful at close range.  It has been widely reported, for example in \cite{gregory1966book}, that beyond about 6 meters, the human visual system is essentially monocular.  But there is mounting evidence that the human stereo system is actually much more capable than that.  Multiple studies have shown metric depth estimation up to 20~meters \cite{levin1993pp,allison2009jov}; and, although error increases as disparity increases \cite{hibbard2017crpi}, controlled experiments have confirmed that scaled disparity can be estimated up to 300~m, even without any depth cues from monocular vision \cite{palmisano2010jov}.  Moreover, since the human visual system is capable of estimating disparity as small as a few seconds of arc \cite{palmisano2010jov}, there is reason to believe that the distance could be 1 km or greater, with some evidence supporting such a claim provided by the experiments of \cite{cormack1984pp}.  Note that an artificial stereo system whose baseline is wider than the average 65~mm interpupillary distance of the human visual system has the potential to provide even greater accuracy.

This question takes on renewed significance in the context of self-driving cars, since most automobile manufacturers and experimental autonomous vehicles do not install stereo cameras in their vehicles.\footnote{To the authors' knowledge, at the time of this writing stereo cameras can be found only on certain models of Mercedes and Subaru vehicles; no major autonomous platform uses them.} Rather, these systems rely on various combinations of monocular cameras, lidar, radar, and sonar sensors.\footnote{Tesla vehicles, for example, are equipped with monocular cameras, sonar, and radar, but no lidar.  Despite having multiple foveated cameras for wider field of view, such vehicles do not rely upon depth from stereopsis.} For detecting static obstacles such as trees, poles, railings, and concrete barriers, most systems rely on cameras and/or lidar.  Although it is beyond the scope of this paper whether monocular cameras are sufficient for self-driving behavior (certainly people with monocular vision can drive safely in most situations), or whether stereo is better than lidar, we argue that the proper engineering approach to such a safety-critical system is to leverage all available sensors rather than assume they are not needed; thus, we believe that it is important to accurately assess the increased error in depth estimation when relying upon monocular cameras.

At typical highway speeds, the braking distance required to completely stop before impact necessitates observing an unforeseen stopped object approximately 100~m away.  Intrigued by the reported success of monocular depth, we tried some recent algorithms, only to discover that monocular depth is not able to achieve accuracies anywhere close to that requirement.  We then turned our attention to stereo, where significant progress has been made in recent years in applying deep learning to the problem \cite{seki2017cvpr:sgmnet,seki2016bmvc:patch,guney2015cvpr:displet,zbontar2016jmlr:stereo,shaked2017cvpr:highway,zhong2017arx:ssl,gidaris2017cvpr:ddr,kendall2017iccv:gcnet,pang2017arx:crl}.  An excellent overview of recent stereo algorithms can be found in \cite{kendall2017iccv:gcnet}.  In this flurry of activity, a variety of architectures have been proposed, but there has been no systematic study as to how these design choices impact quality. One purpose of this paper is thus to investigate several of these options in order to quantify their impact, which we do in Sec.~\ref{sec:experimental}.  In the context of this study, we develop a novel semi-supervised stereo approach, which we present in Sec.~\ref{sec:deepsn}.  First, however, we illustrate the limitations of monocular depth estimation in the next section.

\section{Difficulties of Monocular Depth Estimation}

\begin{figure}
\begin{center}
   \includegraphics[width=1.0\linewidth]{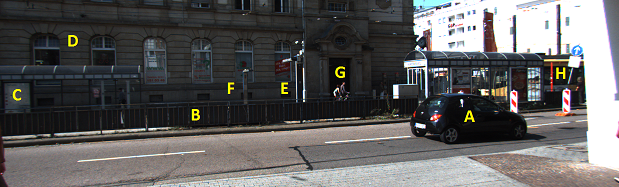}
\end{center}
   \caption{An image from the KITTI dataset \cite{geiger2013ijrr:kitti} showing a road in front of a pair of train tracks in front of a building. Several items of interest are highlighted:  (A) car, (B) fence, (C) depot, (D) building, (E) near pole, (F) far pole, (G) people, and (H) departing train.  The building is 30~m from the camera.}
\label{fig:image_of_monostereo}
\end{figure}

To appreciate the gap between mono and stereo vision, consider the image of Fig.~\ref{fig:image_of_monostereo}, with several points of interest highlighted.  Without knowing the scene, if you were to ask yourself whether the width of the near road (on which the car (A) sits) is greater than the width of the far tracks (distance between the near and far poles (E and F)), you might be tempted to answer in the affirmative.  After all, the road not only occupies more pixels in the image (which is to be expected, since it is closer to the camera), but it occupies orders of magnitude more pixels.  We showed this image to several people in our lab, and they all reached the same conclusion:  the road indeed appears to be significantly wider.  As it turns out, if this image is any indication, people are not very good at estimating metric depth from a single image.\footnote{Specifically, we asked 8 people to estimate the distance to the fence (ground truth 14~m) and the distance to the building (ground truth 30~m).  Their estimates on average were 9.3~m and 12.4~m, respectively.  The distances were therefore underestimated by 34\% and 59\%, respectively, and the distance from the fence to the building was underestimated by 81\%.}

\begin{figure}
\begin{center}
\begin{tabular}{c}
  \includegraphics[width=1.0\linewidth]{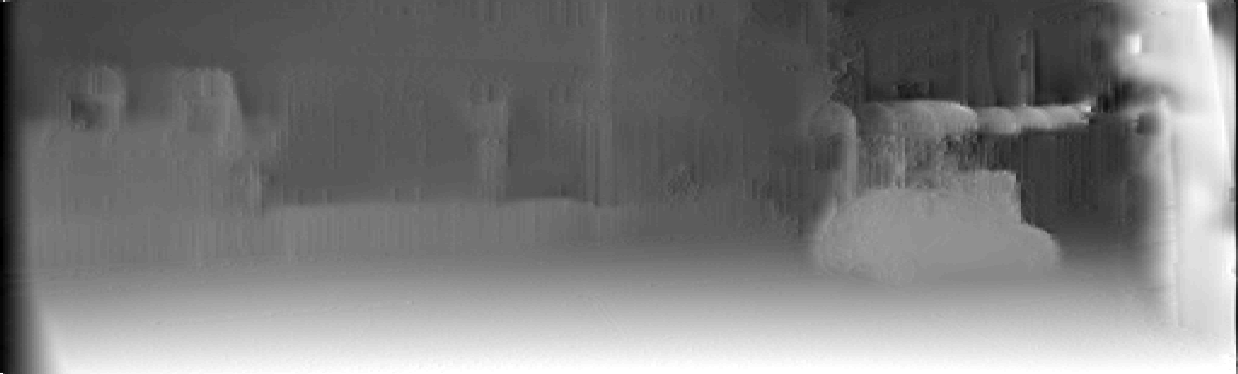} \\
  \includegraphics[width=1.0\linewidth]{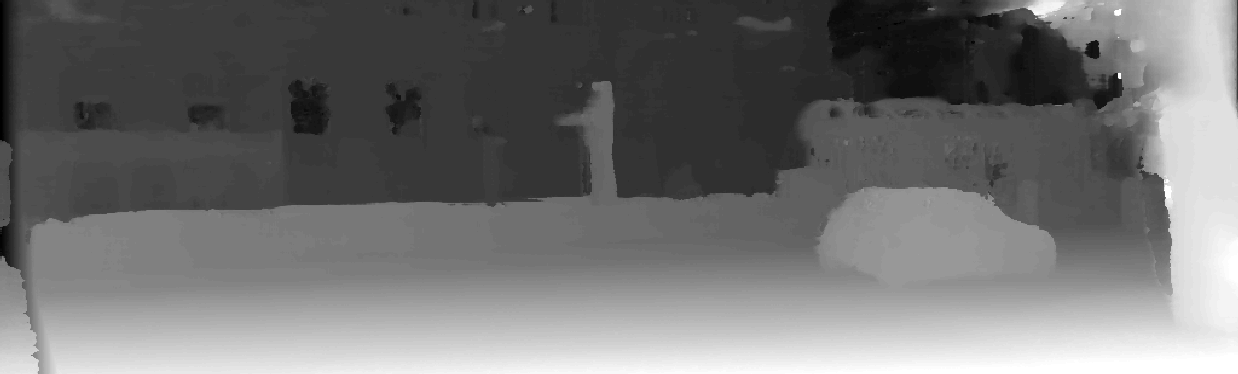}
	\end{tabular}
\end{center}
   \caption{Results of MonoDepth \cite{godard2017cvpr:unsup} (top) vs.\ our stereo algorithm (bottom) on the image (or pair of images, in the latter case) of the previous figure, displayed as depth/disparity maps.}
\label{fig:results_of_monostereo_gray}
\end{figure}

The output of a leading monocular depth algorithm, called MonoDepth \cite{godard2017cvpr:unsup}, is shown in Fig.~\ref{fig:results_of_monostereo_gray},\footnote{Other monocular algorithms produce similar results.} along with the output of our stereo depth algorithm.  At first glance, both results appear plausible.  Although the stereo algorithm preserves crisper object boundaries and appears at least slightly more accurate, it is difficult to tell from the grayscale images just how much the two results differ.  

\begin{figure}
\begin{center}
\begin{tabular}{cc}
  \includegraphics[width=0.47\linewidth]{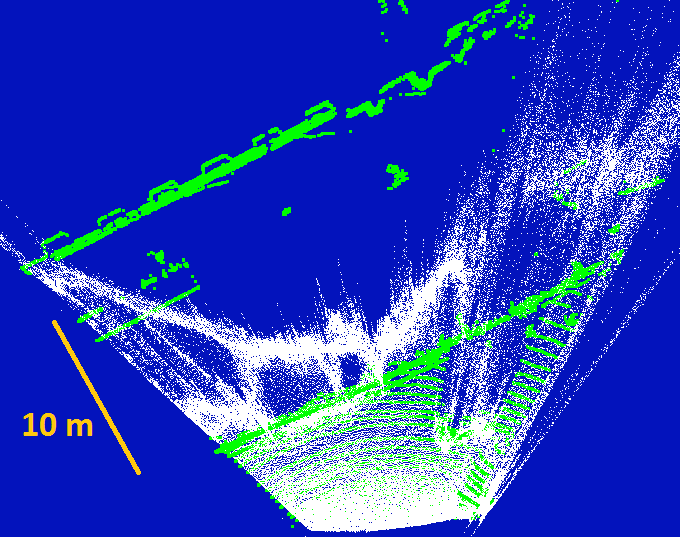} &
	\includegraphics[width=0.47\linewidth]{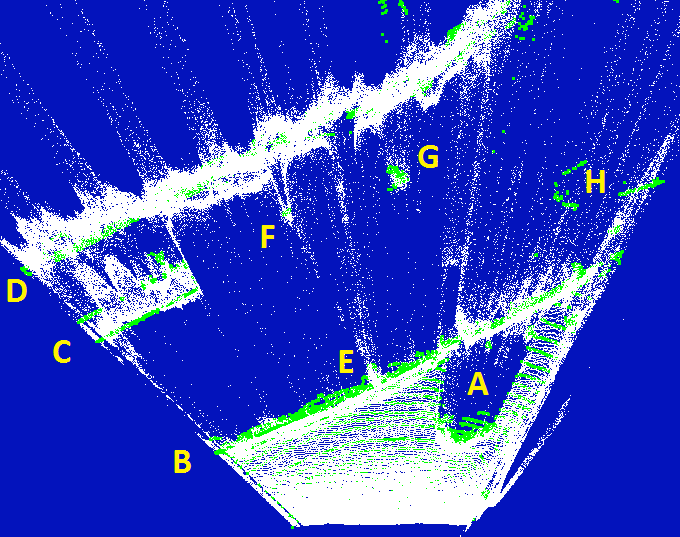} \\
	MonoDepth \cite{godard2017cvpr:unsup} & our stereo algorithm
	\end{tabular}
\end{center}
   \caption{Results of MonoDepth \cite{godard2017cvpr:unsup} (left) vs our stereo algorithm (right), displayed as 3D point clouds from a top-down view.  Green dots indicate ground truth from LIDAR.  The letters indicate objects of interest from Fig.~1.  Note that stereo is able to recover accurate geometry up to at least 30~m away.  (Best viewed in color.)}
\label{fig:results_of_monostereo_topdown}
\end{figure}

In fact, the differences are quite large.  To better appreciate these differences, Fig.~\ref{fig:results_of_monostereo_topdown} shows a top-down view of the point clouds associated with the depth/disparity maps with the ground truth LIDAR data overlaid.  These results reveal that monocular depth is not only inaccurate in an absolute sense (due to the overall scale ambiguity from a single images), it is also inaccurate in recovering details.  In fact, of the 8 objects of interest highlighted in Fig.~1, the monocular algorithm misses nearly all of them---except perhaps the car (A) and some of the fence (B).  In contrast, our stereo algorithm is able to properly detect the car (A), fence (B), depot (C), building (D), near (E) and far (F) poles, and people (G).  The only major object missed by the stereo algorithm is the train (H) leaving the station, which is seen primarily through the transparent depot glass.  These results are even more dramatic when viewed on the screen with freedom to rotate and zoom.  

One could argue that this is not a fair comparison:  obviously stereo is better because it has access to more information.  But that is exactly the point, namely, that stereo algorithms have access to information that monocular algorithms will never have, and such information is crucial for accurately recovering depth.  Therefore, any application that requires accurate depth and can afford to support more than one camera should take advantage of such information.  

To further shed light on this point, notice that the top-down view of the previous figure contains the answer to the question posed at the beginning of the section:  the width of the tracks is approximately the same as that of the road.  Amazingly, the stereo algorithm, with just a single pair of images from a single point in time, is able to recover such information, even though the building behind the tracks is 30~m away.  In contrast, the fact that the human visual system is so easily fooled by the single photograph leads us to believe that the limitation in accuracy for monocular depth is not due to the specific algorithm used but rather is a fundamental hurdle that will prove frustratingly difficult to overcome for a long time.\footnote{Of course, one could use multiple images in time from a single camera to overcome such limitations.  Note, however, that in the context of a self-driving car, the forward direction (which is where information is needed most) is precisely the part of the image containing the least image motion and, hence, the least information.}

\section{Deep Stereo Network}
\label{sec:deepsn}

\begin{figure*}[t]
\begin{center}
   \includegraphics[width=0.9\linewidth]{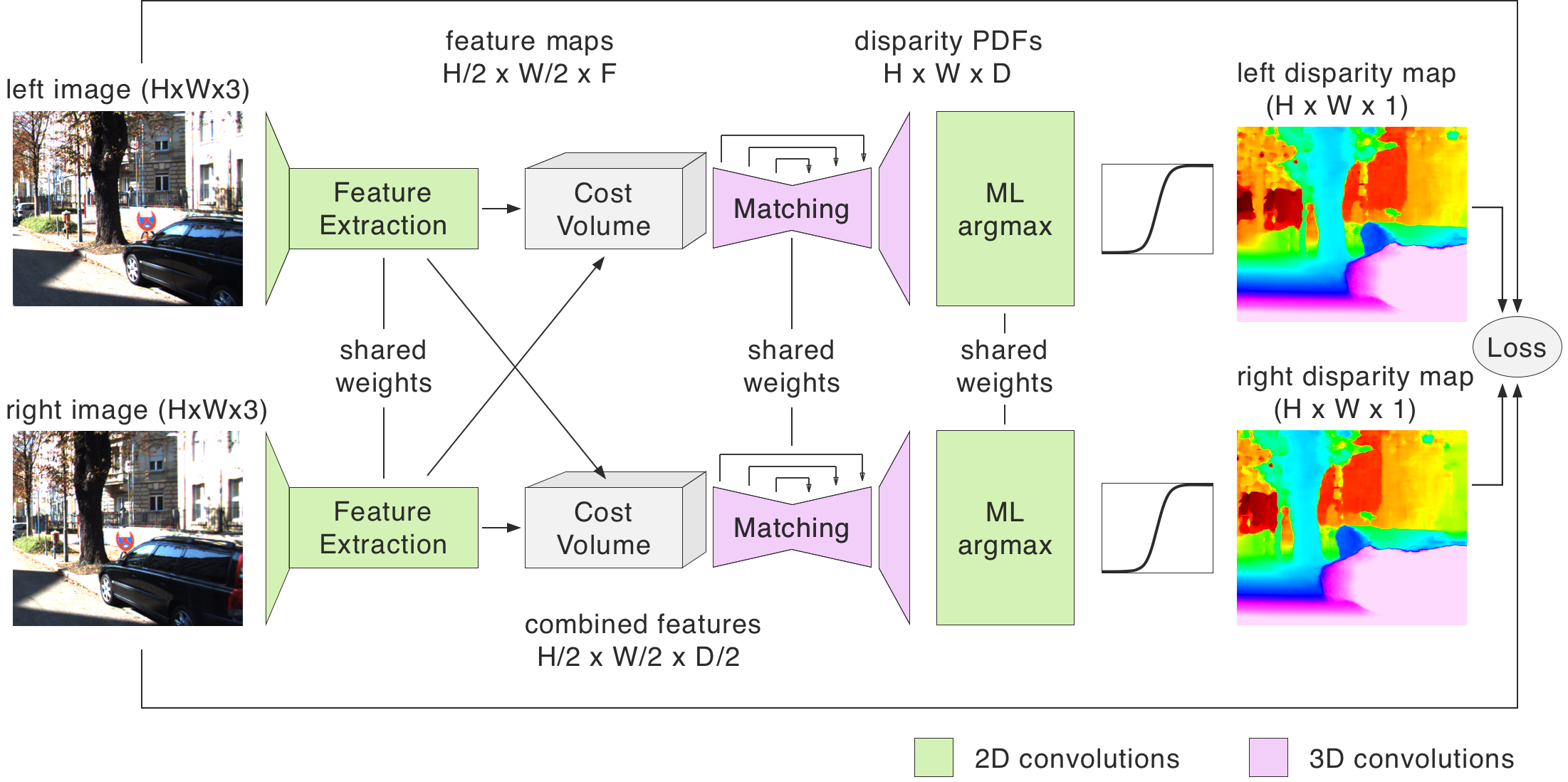}
\end{center}
   \caption{Architecture of our binocular stereo network to estimate disparity (and hence depth).}
\label{fig:networkArch}
\end{figure*}

Recognizing the limitations of monocular depth, we instead use a stereo pair of images.    
Our stereo network, shown in Fig.~\ref{fig:networkArch}, is inspired by the architecture of the recent GC-Net stereo network~\cite{kendall2017iccv:gcnet} which at the time we began the investigation, was the leader of the KITTI 2015 benchmark.
The left and right images (size $H \times W \times C$, where $C=3$ is the number of input channels) are processed by 2D feature extractors based on a residual network architecture that bears resemblance to ResNet-18~\cite{he2016cvpr:resnet}.  The resulting feature tensors (dimensions $\mbox{\sfrac{1}{2}}H \times \mbox{\sfrac{1}{2}}W \times F$, where $F=32$ is the number of features)  are used to create two cost volumes, one for left-right matching and the other for right-left matching. The left-right cost volume is created by sliding the right tensor to the right, along the epipolar lines of the left tensor, up to max disparity.  At corresponding pixel positions, the left and right features are concatenated and copied into the resulting 4D cost volume (dimensions $\mbox{\sfrac{1}{2}}D \times \mbox{\sfrac{1}{2}}H \times \mbox{\sfrac{1}{2}}W \times 2F$, where $D$ is the max disparity). The right-left cost volume is created by repeating this procedure in the opposite direction by sliding the left tensor to the left, along the epipolar lines of the right tensor, also up to the max disparity.  Note that, as in \cite{kendall2017iccv:gcnet}, the first layer of the network downsamples by a factor of two in each direction to reduce both computation and memory in the cost volumes.

These two cost volumes are used in a 3D convolution / deconvolution bottleneck that performs stereo matching by comparing features.  This bottleneck contains a multiscale encoder to perform matching at multiple resolutions, followed by a decoder with skip connections to incorporate information from the various resolutions.  Just as in the feature extraction layers above, the weights in the left and right bottleneck matching units are shared and learned together.  After the last decoder layer, upsampling is used to produce both a left and right tensor (dimensions $D \times H \times W \times 1$) containing matching costs between pixels in the two images.  

At this point it would be natural to apply differentiable soft argmax~\cite{kendall2017iccv:gcnet} to these matching costs (after first converting to probabilities) to determine the best disparity for each pixel.  Soft argmax has the drawback, however, of assuming that all context has already been taken into account, which may not be the case.  To overcome this limitation, we implement a machine-learned argmax (ML-argmax) function using a sequence of 2D convolutions to produce a single value for each pixel which, after passing through a sigmoid, becomes the disparity estimate for that pixel.  We found the sigmoid to be a crucial detail, without which the disparities were not learned correctly.  Our machine-learned argmax is not only able to extract disparities from the disparity PDF tensor, but it is also better at handling uniform or multimodal probability distributions than soft argmax.  Moreover, it yields more stable convergence during training.

Three key differences of our architecture with respect to GC-Net \cite{kendall2017iccv:gcnet} are the following: 1) our semi-supervised loss function which includes both supervised and unsupervised terms, as explained in more detail below; 2) our use of ELU activations \cite{clevert2016iclr:elu} rather than ReLU-batchnorm, which enables the network to train and run faster by obviating the extra operations required by batchnorm; and 3) our novel machine-learned argmax function rather than soft argmax, which allows the network to better incorporate context before making a decision.

\begin{table}%
\begin{center}
\caption{Previous stereo methods have used either supervised or unsupervised training, whereas we use both (semi-supervised).}
\label{tab:supunsup}
{\small
\begin{tabular}{c|c|c}
method & supervised & unsupervised \\
\hline
SGM-Net \cite{seki2017cvpr:sgmnet} & S& $\cdot$  \\
PBCP \cite{seki2016bmvc:patch} & S& $\cdot$  \\
L-ResMatch \cite{shaked2017cvpr:highway} & S& $\cdot$  \\
SsSMnet \cite{zhong2017arx:ssl} & $\cdot$ & U \\
GC-Net \cite{kendall2017iccv:gcnet} & S & $\cdot$ \\
CRL \cite{pang2017arx:crl} & S & $\cdot$ \\
Ours & S & U
\end{tabular}
}
\end{center}
\end{table}

To train the network, we use the following loss function, which combines the supervised term ($E_{\id{lidar}}$) used by most other stereo algorithms 
\cite{seki2017cvpr:sgmnet,seki2016bmvc:patch,shaked2017cvpr:highway,kendall2017iccv:gcnet,pang2017arx:crl}
along with unsupervised terms similar to those used by MonoDepth \cite{godard2017cvpr:unsup}:
\begin{equation}
L = \lambda_1E_{\id{image}} + \lambda_2E_{\id{lidar}} + \lambda_3E_{\id{lr}} + \lambda_4E_{\id{ds}},
\label{eq:loss}
\end{equation}
where 
\begin{eqnarray}
E_{\id{image}} &=& E^l_{\id{image}} + E^r_{\id{image}} \label{eq:myeq2} \\
E_{\id{lidar}} &=& |d_l-{\bar d}_l| + |d_r-{\bar d}_r| \label{eq:lidar} \\
E_{\id{lr}} &=& \frac{1}{n}\sum_{ij}|d_{ij}^l-{\tilde d}_{ij}^l| + \frac{1}{n}\sum_{ij}|d_{ij}^r-{\tilde d}_{ij}^r| \label{eq:myeq4} \\
E_{\id{ds}} &=& E^l_{\id{ds}} + E^r_{\id{ds}}. \label{eq:myeq5}
\label{eq:loss2}
\end{eqnarray}
Note that Eq.~(\ref{eq:myeq2}) ensures photometric consistency, Eq.~(\ref{eq:lidar}) compares the estimated disparities to the sparse LIDAR data, Eq.~(\ref{eq:myeq4}) ensures that the left and right disparity maps are consistent with each other, and Eq.~(\ref{eq:myeq5}) encourages the disparity maps to be piecewise smooth, respectively, where
\begin{eqnarray*}
E^l_{\id{image}} &=& \frac{1}{n} \sum_{i,j} \alpha \frac{1-\id{SSIM}(I_{ij}^l,{\tilde I}_{ij}^l)}{2} 
	+ (1-\alpha) |I_{ij}^l - {\tilde I}_{ij}^l| \\
E^l_{\id{ds}} &=& \frac{1}{n} \sum_{i,j} |\partial_x d_{ij}^l | e^{-\| \partial_x I_{i,j}^l\|} 
		+ |\partial_y d_{ij}^l | e^{-\| \partial_y I_{i,j}^l\|} 
\label{eq:loss3}
\end{eqnarray*}
and similarly for $E^r_{\id{image}}$ and $E^r_{\id{ds}}$.  The quantities above are defined as
\begin{eqnarray}
{\tilde I}^l &=& w_{rl}(I_r,d_l) \\
{\tilde I}^r &=& w_{lr}(I_l,d_r) \\
{\tilde d}^l &=& w_{rl}(d_r,d_l) \\
{\tilde d}^r &=& w_{lr}(d_l,d_r) \\
w_{lr}(I,d) &=& (x,y) \mapsto I(x-d(x,y),y) \label{eq:wlr} \\
w_{rl}(I,d) &=& (x,y) \mapsto I(x+d(x,y),y) \label{eq:wrl} \\
\id{SSIM}(x,y) &=& \!\!\!\!\left(\frac{2\mu_x\mu_y+c_1}{\mu_x^2+\mu_y^2+c_1}\right)
							\!\!\left(\frac{2\sigma_{xy}+c_2}{\sigma_x^2+\sigma_y^2+c_2}\right)
\label{eq:loss4}
\end{eqnarray}
Note that $I_l$ and $I_r$ are the input images, $d_l$ and $d_r$ are the estimated disparity maps output by the network, ${\bar d}_l$ and ${\bar d}_r$ are the ground truth disparity maps obtained from LIDAR, SSIM is the structural similarity index \cite{zhou2004itip:ssim,zhao2017itci:ssim,godard2017cvpr:unsup}, $n$ is the number of pixels, and $c_1=10^{-4}$ and $c_2=10^{-3}$ are constants to avoid dividing by zero.  Note that in Eqs.~\eqref{eq:wlr}--\eqref{eq:wrl} the coordinates are often non-integers, in which case we use bilinear interpolation, implemented similar to \cite{jaderberg2015nips:stn}.

\section{Experimental Results}
\label{sec:experimental}

To evaluate our network as well as its variants, we trained and tested on the KITTI dataset \cite{geiger2013ijrr:kitti}, requiring more than 40 GPU-days.  For training, we used the 29K training images\footnote{This the same training set split used by MonoDepth \cite{godard2017cvpr:unsup}.} with sparse LIDAR for ground truth.  To our knowledge, we are the first to combine supervised and unsupervised learning for training a deep stereo network, see Tab.~\ref{tab:supunsup}.  The network was implemented in TensorFlow and trained for 85000 iterations (approx.~2.9 epochs) with a batch size of 1 on an NVIDIA Titan X GPU.  We use the Adam optimizer starting with a learning rate of $10^{-4}$, which was reduced over time.  We then tested the network on the 200 training images from the KITTI 2015 benchmark, which contain sparse LIDAR ground truth augmented by dense depth on some vehicles from fitted 3D CAD models.  (Note that this process separates the training and testing datasets, since the 200 images are from 33 scenes that are distinct from the 28 scenes associated with the 29K training images.)  Like other authors, we used these 200 training images for testing, since the ground truth for the KITTI 2015 test images is not publicly available, and submission to the website is limited.  

\begin{table*}%
\caption{Stereo architecture variants.  The top row describes our deep stereo network, the second row is our baseline system (without machine-learned argmax), and the remaining rows describe variations of the baseline.  Feature extraction is identical in all cases except for ``small / tiny''.  The cost volume is constructed using either concatenation or correlation of features, leading to either a 4D or 3D cost volume, respectively; there are actually two cost volumes except for ``single tower''. The bottleneck layers are smaller in ``small / tiny'' and replaced by convolutional layers in ``no bottleneck''; ``tiny'' has half as many 3D filters as ``small'' in the bottleneck. The aggregator is soft argmax except for our network, which uses our machine-learned argmax.  For the layer notation, see the text.}
\label{tab:diffArch}
\begin{center}
{\small
\begin{tabular}{l|c|l|c|c|c}
model & features & cost volume & bottleneck & upsampler & aggregator \\
& (2D conv.) & & (3D conv./deconv.) & (3D deconv.) & (2D conv.) \\
\hline
ML-argmax (ours) & $(\onelessspace \downarrow_1, 8(2C_+), 1C)$ & concat. (4D) & $(\lessspace{4}\downarrow_3,2C,\lessspace{4}\uparrow_{1+})$ & $\onelessspace \uparrow_1$ & ML-argmax (5C) \\
baseline (ours) & $(\onelessspace \downarrow_1, 8(2C_+), 1C)$ & concat. (4D) & $(\lessspace{4}\downarrow_3,2C,\lessspace{4}\uparrow_{1+})$ & $\onelessspace \uparrow_1$ & soft-argmax \\
correlation & $(\onelessspace \downarrow_1, 8(2C_+), 1C)$ & correlation (3D) & $(\lessspace{4}\downarrow_3,2C,\lessspace{4}\uparrow_{1+})$ & $\onelessspace \uparrow_1$ & soft-argmax \\
no bottleneck & $(\onelessspace \downarrow_1, 8(2C_+), 1C)$ & concat. (4D) & $(2C)$ & $\onelessspace \uparrow_1$ & soft-argmax \\
single tower & $(\onelessspace \downarrow_1, 8(2C_+), 1C)$ & concat. (4D, single) & $(\lessspace{4}\downarrow_3,2C,\lessspace{4}\uparrow_{1+})$ & $\onelessspace \uparrow_1$ & soft-argmax \\
small / tiny & $(5C)$ & concat. (4D) & $(\lessspace{2}\downarrow_3,2C,\lessspace{2}\uparrow_{1+})$ & $\onelessspace \uparrow_1$ & soft-argmax \\
\end{tabular}
}
\end{center}
\end{table*}

For all tests, the input images (which are originally different sizes) were resized to $1024 \times 320$; and for the LIDAR-only experiments the images were further cropped to remove $37.8$\% of the upper part.  The maximum disparity was set to $D=96$.  No scaling was done on the input images, but the LIDAR values were scaled to be between 0 and 1.  The same procedure was used for all variants, and no postprocessing was done on the results.

The various architectures that we tested are listed in Tab.~\ref{tab:diffArch}.  These variants are named with respect to a baseline architecture.  Thus, our ML-argmax network is an extension to the baseline, whereas the other variants are less powerful versions that either replace concatenation with cross-correlation (sliding dot product), replace the bottleneck layers with simpler convolutional layers, remove one of the two towers, or use a smaller number of weights.\footnote{We also tried replacing 3D convolutions with 2D convolutions (similar to \cite{mayer2016cvpr:dispnet}), but the network never converged.}  The single-tower version has a modified loss function with all terms involving the right disparity map removed.

The notation of the layers in the table is as follows:  $mB_k$ means $m$ blocks of type $B$ with $k$ layers in the block.  Thus, $\onelessspace \downarrow_1$ means a single downsampling layer, $\onelessspace \uparrow_1$ means a single upsampling layer, and $2C$ means two convolutional layers.  The subscript $_+$ indicates a residual connection, so $8(2C_+)$ means 8 superblocks, where each superblock consists of 2 blocks of single convolutional layers accepting residual connections.

\begin{figure}[t]
\begin{center}
	\begin{tabular}{c}
   \hspace{-3ex} \includegraphics[width=0.9\linewidth]{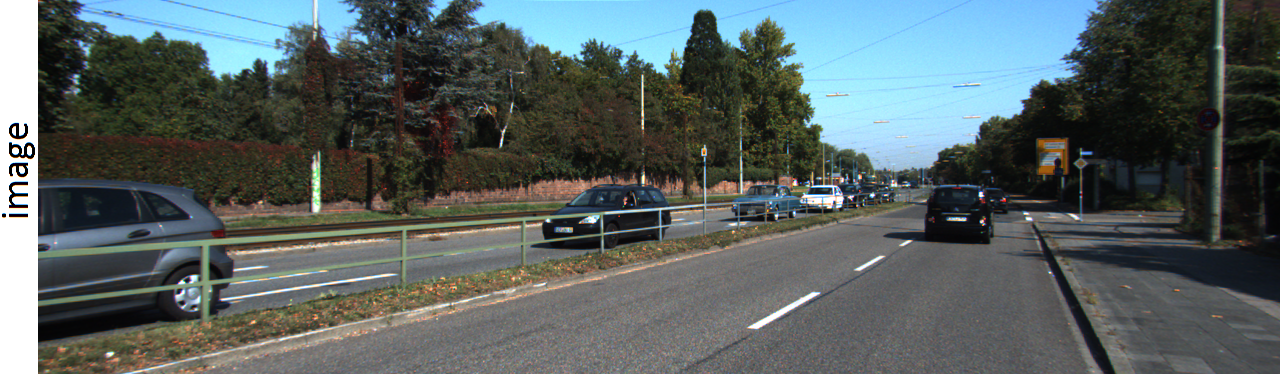} \\
   \hspace{-3ex} \includegraphics[width=0.9\linewidth]{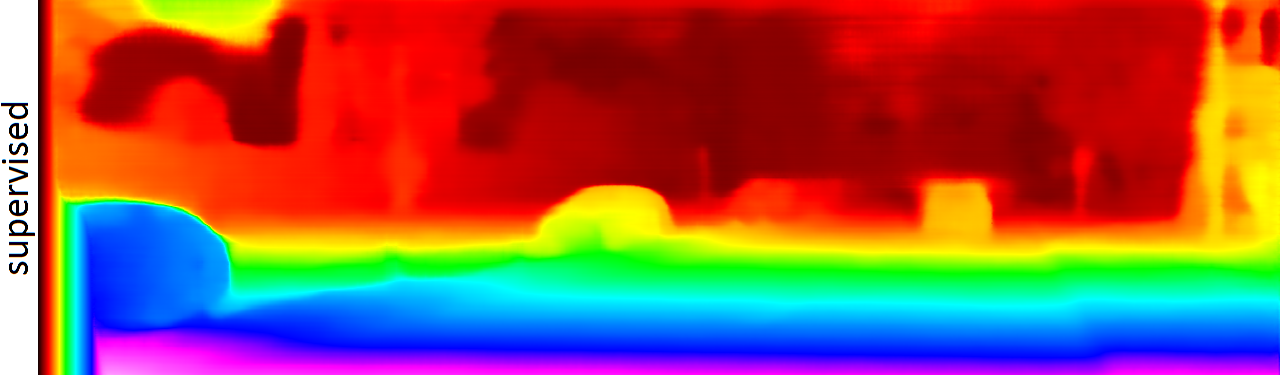} \\
   \hspace{-3ex} \includegraphics[width=0.9\linewidth]{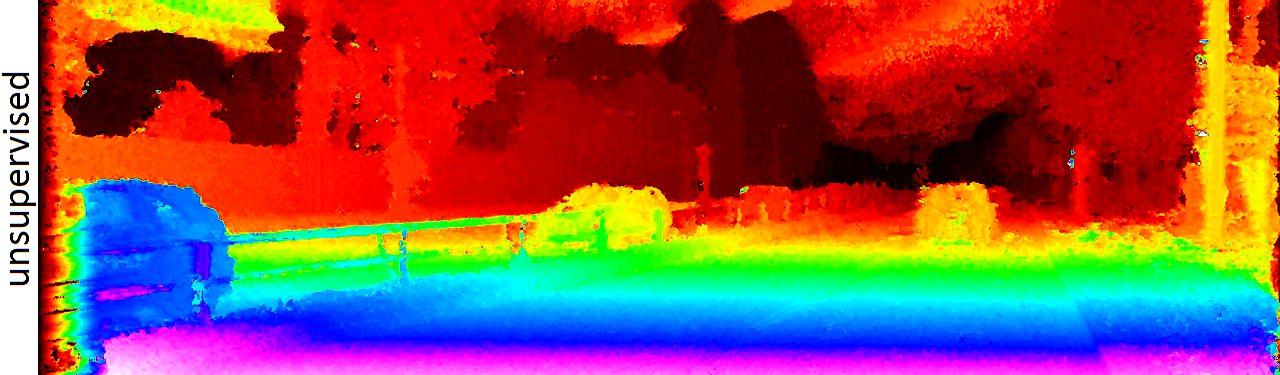} \\
   \hspace{-3ex} \includegraphics[width=0.9\linewidth]{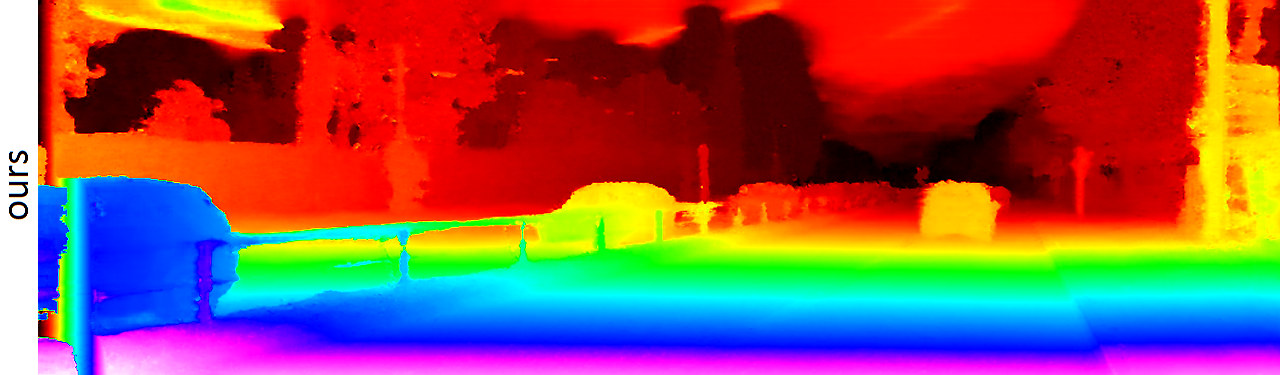} \\
	 \includegraphics[width=0.97\linewidth]{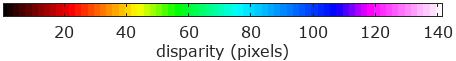}
	\end{tabular}
\end{center}
   \caption{From top to bottom:  an image, and results from supervised (LIDAR), unsupervised (photometric consistency), and semi-supervised (both) learning.  Notice that the sparse LIDAR data leads to smoothed results that misses fine details (e.g., the fence), and the photometric loss recovers fine details but yields noisy results.  Our semi-supervised approach combines the best of both.  See the text for an explanation of the colormap.}
\label{fig:qualitative1}
\end{figure}

Our first set of experiments was aimed at comparing unsupervised, supervised, and semi-supervised learning.  The results of three variant architectures, along with monocular depth, are shown in Tab.~\ref{tab:ressemisup}, which contains the D1-all error of all pixels as defined by KITTI (the percentage of pixels with an error at least 3 disparity levels or at least 5\%).  This error is the percentage of outliers.  Surprisingly, in all cases the unsupervised (photometric) loss yielded better results than the supervised loss (LIDAR).  The best results were obtained by combining the two, because photometric and LIDAR data complement each other:  LIDAR is accurate at all depths, but its sparsity leads to blurrier results, and it misses the fine structure, whereas photometric consistency allows the network to recover fine-grained surfaces but suffers from loss in accuracy as depth increases.  These observations are clearly seen in the example of Fig.~\ref{fig:qualitative1}.

MonoDepth \cite{godard2017cvpr:unsup} performed noticeably worse, thus demonstrating (as explained earlier) that the gap between mono and stereo is significant.  (We used MonoDepth because it is a leading monocular depth algorithm whose code is available online; other monocular algorithms perform similarly.)  Note that only the relative values are important here; the absolute values are large in general from testing on images with dense ground truth despite being trained only on images with sparse ground truth.  For these experiments as well as the next, the relative weights in the loss function were set to $\lambda_1=\lambda_3=1.0$ for lidar and for photo, or $\lambda_1=0.01$, $\lambda_3=0.1$ for lidar+photo; and $\lambda_4=0.1$, $\alpha=0.85$.

\begin{table}%
\caption{Improvement from combining supervised (LIDAR) with unsupervised (photometric consistency) learning.  Shown are D1-all errors on the 200 KITTI 2015 augmented training images after training on 29K KITTI images with sparse ground truth.  Note that only relative values are meaningful; see text.}
\label{tab:ressemisup}
\begin{center}
{\small
\begin{tabular}{l|c|c|c}
model & lidar & photo & lidar+photo \\
\hline
MonoDepth \cite{godard2017cvpr:unsup} & - & 32.8\% & - \\
no bottleneck & 21.3\% & 18.6\% & 14.5\% \\
correlation & 14.6\% & 13.3\% & 12.9\% \\
baseline (ours) & 15.0\% & 12.9\% & 8.8\% 
\end{tabular}
}
\end{center}
\end{table}

Having established the benefit of combining supervised and unsupervised learning, the second set of experiments aimed at providing further comparison among the architecture variants.  Results are shown in Tab.~\ref{tab:resnac}.  A significant improvement is achieved by our machine-learned argmax.  Somewhat surprisingly, reducing the size of the network substantially by either using a smaller network, cross-correlation, or removing one of the towers entirely has only a slight effect on error, despite the fact that a single tower requires 1.8X less memory, cross-correlation requires 64X less memory, the small network contains 36\% fewer weights, and the tiny network contains 82\% fewer weights.  From these data we also see that the bottleneck is extremely important to extract information from the cost volume, and that concatenation is noticeably better than correlation, thus confirming the claim of \cite{kendall2017iccv:gcnet}.  

\begin{table}%
\caption{Influence of various network architecture changes.  Shown are D1-all errors on the 200 KITTI 2015 augmented training images after training on 29K KITTI images with sparse ground truth.  Network size is measured by the number of weights.  Note that only relative values are meaningful; see text.}
\label{tab:resnac}
\begin{center}
{\small
\begin{tabular}{l|c|c}
model & size & lidar+photo \\
\hline
no bottleneck & 0.2M & 14.5\% \\
correlation & 2.7M & 12.9\% \\
small & 1.8M & 9.8\% \\
tiny & 0.5M & 11.9\% \\
single tower & 2.8M & 10.1\% \\
baseline (ours) & 2.8M & 8.8\% \\
ML-argmax (ours) & 3.1M & 8.7\% \\
\end{tabular}
}
\end{center}
\end{table}

\begin{table*}%
\caption{Results of our network compared with the leaders of the KITTI 2015 website, as of 2018-Mar-19.  Anonymous results are excluded.  With fine-tuning, our network achieves errors that are competitive with state-of-the-art, even without training on synthetic data.}
\label{tab:kittiresults}
\begin{center}
{\small
\begin{tabular}{c||c|c|c||c|c|c}
		& \multicolumn{3}{c||}{Non-occluded} & \multicolumn{3}{c}{All} \\
model & D1-bg & D1-fg & D1-all  & D1-bg & D1-fg & D1-all \\
\hline
DispNetC \cite{mayer2016cvpr:dispnet} & 4.1\% & 3.7\% & 4.1\% & 4.3\% & 4.4\% & 4.3\% \\
SGM-Net \cite{seki2017cvpr:sgmnet} & 2.2\% & 7.4\% & 3.1\% & 2.7\% & 8.6\% & 3.7\% \\ 
PBCP \cite{seki2016bmvc:patch} & 2.3\% & 7.7\% & 3.2\% & 2.6\% & 8.7\% & 3.6\% \\
Displets v2 \cite{guney2015cvpr:displet} & 2.7\% & 5.0\% & 3.1\% & 3.0\% & 5.6\% & 3.4\% \\
L-ResMatch \cite{shaked2017cvpr:highway} & 2.4\% & 5.7\% & 2.9\% & 2.7\% & 7.0\% & 3.4\% \\
SsSMnet \cite{zhong2017arx:ssl} & 2.5\% & 6.1\% & 3.0\% & 2.7\% & 6.9\% & 3.4\% \\
DRR \cite{gidaris2017cvpr:ddr} & 2.3\% & 4.9\% & 2.8\% & 2.6\% & 6.0\% & 3.2\% \\
GC-Net \cite{kendall2017iccv:gcnet} & 2.0\% & 5.6\% & 2.6\% & 2.2\% & 6.2\% & 2.9\% \\
CRL \cite{pang2017arx:crl} & 2.3\% & 3.1\% & 2.5\% & 2.5\% & 3.6\% & 2.7\% \\
iResNet \cite{liang2017arx:stereo} & 2.1\% & 2.8\% & 2.2\% & 2.3\% & 3.4\% & 2.4\% \\
\hline
Ours (no fine-tuning) & 2.7\% & 13.6\% & 4.5\% & 3.2\% & 14.8\% & 5.1\% \\
Ours (fine-tuned) & 2.1\% & 4.5\% & 2.5\% & 2.7\% & 6.0\% & 3.2\%
\end{tabular}
}
\end{center}
\end{table*}

To test on the official KITTI 2015 benchmark,\footnote{\url{http://www.cvlibs.net/datasets/kitti}} we submitted two versions.  The first version is exactly the same baseline network as described above without retraining or fine-tuning, except that we validated using the 200 KITTI training images to learn the relative weights, $\lambda_1=0.01$, $\lambda_2=1.0$, $\lambda_3=\lambda_4=0.1$; and we set the maximum disparity to $D=136$.  The results, shown in Tab.~\ref{tab:kittiresults}, are significantly better (due to this reweighting) than on the augmented training images, achieving 5.1\% D1-all error on all pixels.  Although this is not competitive with recent techniques, it is surprisingly good considering that the network was not trained on dense data.  For the next result, we took this same network and fine-tuned it using the 200 KITTI 2015 augmented training images.  After fine-tuning, our results are competitive with state-of-the-art, achieving 3.2\% D1-all error on all pixels and only 2.5\% on non-occluded pixels.  These results were achieved without any postprocessing of the data.

Our baseline network achieves results similar to those of GC-Net \cite{kendall2017iccv:gcnet}, actually winning on three of the six metrics.  The remaining difference between the results is likely due to GC-Net's pretraining on dense data from the Scene Flow dataset \cite{mayer2016cvpr:dispnet}.  As a result, our network performs less well around the boundaries of objects, since it has seen very little dense ground truth data.  Similar arguments can be made for other competing algorithms, such as CRL \cite{pang2017arx:crl} and iResNet \cite{liang2017arx:stereo}.  However, the focus of this paper was to examine the influence of network architecture and loss functions rather than datasets.  It would be worthwhile in the future to also study the influence of training and pretraining datasets, as well as the use of synthetic and real data.

Fig.~\ref{fig:qualitative1} highlights an advantage of our approach over GC-Net and other supervised approaches.  Because our network is trained in a semi-supervised manner, it is able to recover fine detail, such as the fence rails and posts.  The sparse LIDAR data in the KITTI dataset rarely captures this detail, as seen in the second row of the figure.  As a result, all stereo algorithms trained on sparse LIDAR only (including GC-Net) will miss this important structure.  However, since the LIDAR on which the KITTI ground truth is based often misses such detail itself, algorithms (such as ours) are not rewarded by the KITTI 2015 stereo benchmark metrics for correctly recovering the detail.

\begin{figure*}
   \includegraphics[width=1.0\linewidth]{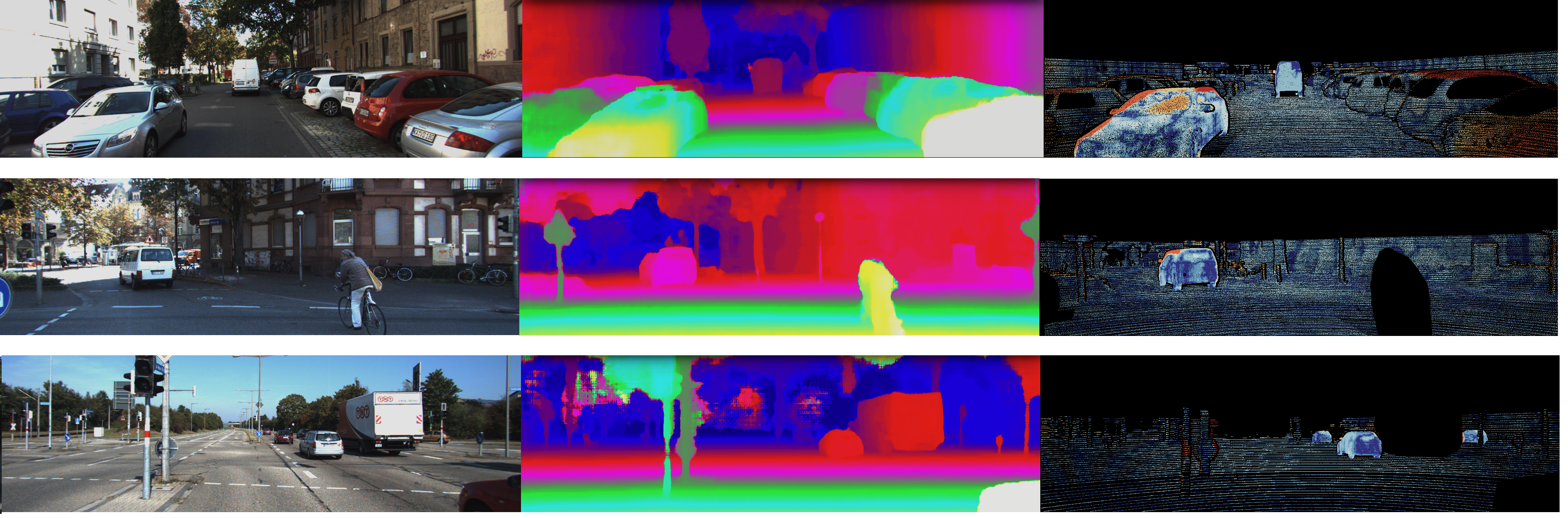}
   \caption{Example results of our algorithm on the KITTI 2015 testing dataset, from the KITTI website.  From left to right:  left input image, disparity map, and error, using the KITTI color maps.}
\label{fig:additionalkitti}
\end{figure*}

The colormap used in Fig.~\ref{fig:qualitative1} was generated by traversing the vertices of the RGB cube in the order KRYGCBMW, which uniquely ensures a Hamming distance of 1 between consecutive vertices (to avoid blending artifacts) and preserves the order of the rainbow.  Distances are scaled so that $\Delta_E$ according to CIE1976 is the same between consecutive vertices.  All images are scaled in the same way, thus preserving the color to disparity mapping.  Objections to rainbow color maps \cite{kovesi2015arx:color,borland2007icga:rainbow} do not appear relevant to structured data such as disparity maps.

Additional results of our final fine-tuned network on the KITTI 2015 online testing dataset are shown in 
Fig.~\ref{fig:additionalkitti}, using the KITTI color maps.  Note that the algorithm accurately detects vehicles, cyclists, buildings, trees, and poles, in addition to the road plane.  In particular, notice in the third row that the interior of the white truck is estimated properly despite the lack of texture.

Tab.~\ref{tab:runtimes} shows the computation time of the various models on different architectures.  Note that with our custom runtime (based on TensorRT / cuDNN), we are able to achieve near real-time performance (almost 20 fps) on Titan XP, as well as efficient performance on the embedded Jetson TX2.\footnote{Our custom runtime, which implements a set of custom plugins for Tensor RT that implement 3D convolutions / deconvolutions, cost volume creation, soft argmax, and ELU, is available at \url{https://github.com/NVIDIA-Jetson/redtail}.}  As far as we know, this is the first deep-learning stereo network ported to embedded hardware.

\begin{table}%
\begin{center}
\caption{Computation time (milliseconds) for different stereo models on various GPU architectures (NVIDIA Titan XP, GTX 1060, and Jetson TX2).  Resolution shows the image dimensions and max disparity, TF indicates TensorFlow runtime, \emph{opt} indicates our custom runtime based on TensorRT / cuDNN, and \emph{OOM} indicates ``out of memory'' exception.  Note that our runtime is necessary for Jetson TX2 because TensorFlow does not run on that board.}
\label{tab:runtimes}
{\small
\begin{tabular}{c|c||c|c||c|c||c}
& & \multicolumn{2}{c||}{Titan XP} & \multicolumn{2}{c||}{GTX 1060} & \multicolumn{1}{c}{TX2} \\
\hline
& resolution & TF & opt & TF& opt& opt\\
\hline
baseline & {\scriptsize 1025\littlex 321\littlex 136} & 950 & 650 & \emph{OOM} & 1900 & 11000 \\
small & {\scriptsize 1025\littlex 321\littlex 96} & 800 & 450 & 2500 & 1150 & 7800 \\
small & {\scriptsize 513\littlex 161\littlex 48} & 280 & 170 & 550 & 300 & 990 \\
tiny & {\scriptsize 513\littlex 161\littlex 48} & 75 & 42 & 120 & 64 & 370 
\end{tabular}
}
\end{center}
\end{table}

\section{Conclusion}

We have shown that a significant gap exists between monocular and stereo depth estimation.  We also presented a careful analysis of various deep-learning-based stereo neural network architectures and loss functions.  Based on this analysis, we propose a novel approach combining a cost volume with concatenated features, 3D convolutions for matching, and machine-learned argmax for disparity extraction, trained in a semi-supervised manner that combines LIDAR and photometric data.  We show competitive results on the standard KITTI 2015 stereo benchmark, as well as superior ability to extract fine details when compared with approaches trained using only LIDAR.  Future work should be aimed at real-time performance, detecting objects at infinity (e.g., skies), and handling occlusions.

\section{Acknowledgments}

We thank Vijay Baiyya and Thang To for their assistance, and Nicholas Haemel for supporting this research.

{\small
\bibliographystyle{ieee}
\bibliography{cvpr2018-stereo}

\begin{thebibliography}{10}\itemsep=-1pt

\bibitem{allison2009jov}
R.~S. Allison, B.~J. Gillam, and E.~Vecellio.
\newblock Binocular depth discrimination and estimation beyond interaction
  space.
\newblock {\em Journal of Vision}, 9(1), Jan. 2009.

\bibitem{borland2007icga:rainbow}
D.~Borland and R.~M. {Taylor II}.
\newblock Rainbow color map (still) considered harmful.
\newblock {\em IEEE Computer Graphics and Applications}, 27(2), Mar. 2007.

\bibitem{clevert2016iclr:elu}
D.-A. Clevert, T.~Unterthiner, and S.~Hochreiter.
\newblock Fast and accurate deep network learning by exponential linear units
  ({ELU}s).
\newblock In {\em ICLR}, 2016.

\bibitem{cormack1984pp}
R.~H. Cormack.
\newblock Stereoscopic depth perception at far observation distances.
\newblock {\em Perception \& Psychophysics}, 35(5):423–428, Sept. 1984.

\bibitem{eigen2014nips:mono}
D.~Eigen, C.~Puhrsch, and R.~Fergus.
\newblock Depth map prediction from a single image using a multi-scale deep
  network.
\newblock In {\em NIPS}, 2014.

\bibitem{garg2016eccv:cnn}
R.~Garg, V.~K. {BG}, G.~Carneiro, and I.~Reid.
\newblock Unsupervised {CNN} for single view depth estimation: {G}eometry to
  the rescue.
\newblock In {\em ECCV}, 2016.

\bibitem{geiger2013ijrr:kitti}
A.~Geiger, P.~Lenz, C.~Stiller, and R.~Urtasun.
\newblock Vision meets robotics: {T}he {KITTI} dataset.
\newblock {\em IJRR}, 32(11):1231--1237, Sept. 2013.

\bibitem{gidaris2017cvpr:ddr}
S.~Gidaris and N.~Komodakis.
\newblock Detect, replace, refine: {D}eep structured prediction for pixel wise
  labeling.
\newblock In {\em CVPR}, 2017.

\bibitem{godard2017cvpr:unsup}
C.~Godard, O.~M. Aodha, and G.~J. Brostow.
\newblock Unsupervised monocular depth estimation with left-right consistency.
\newblock In {\em CVPR}, 2017.

\bibitem{gregory1966book}
R.~L. Gregory.
\newblock {\em Eye and brain}.
\newblock London: World University Library, 1966.

\bibitem{guney2015cvpr:displet}
F.~Guney and A.~Geiger.
\newblock Displets: {R}esolving stereo ambiguities using object knowledge.
\newblock In {\em CVPR}, 2015.

\bibitem{he2016cvpr:resnet}
K.~He, X.~Zhang, S.~Ren, and J.~Sun.
\newblock Deep residual learning for image recognition.
\newblock In {\em CVPR}, 2016.

\bibitem{hibbard2017crpi}
P.~B. Hibbard, A.~E. Haines, and R.~L. Hornsey.
\newblock Magnitude, precision, and realism of depth perception in stereoscopic
  vision.
\newblock {\em Cognitive Research: Principles and Implications}, 2(1):25, 2017.

\bibitem{jaderberg2015nips:stn}
M.~Jaderberg, K.~Simonyan, A.~Zisserman, and K.~Kavukcuoglu.
\newblock Spatial transformer networks.
\newblock In {\em NIPS}, 2015.

\bibitem{kendall2017iccv:gcnet}
A.~Kendall, H.~Martirosyan, S.~Dasgupta, P.~Henry, R.~Kennedy, A.~Bachrach, and
  A.~Bry.
\newblock End-to-end learning of geometry and context for deep stereo
  regression.
\newblock In {\em ICCV}, 2017.

\bibitem{kovesi2015arx:color}
P.~Kovesi.
\newblock Good color maps: {H}ow to design them.
\newblock In {\em arXiv:1509.03700}, 2015.

\bibitem{kuznietsov2017cvpr:mono}
Y.~Kuznietsov, J.~St{\"u}ckler, and B.~Leibe.
\newblock Semi-supervised deep learning for monocular depth map prediction.
\newblock In {\em CVPR}, 2017.

\bibitem{levin1993pp}
C.~A. Levin and R.~N. Haber.
\newblock Visual angle as a determinant of perceived interobject distance.
\newblock {\em Perception \& Psychophysics}, 54(2):250--259, Mar. 1993.

\bibitem{liang2017arx:stereo}
Z.~Liang, Y.~Feng, Y.~Guo, H.~Liu, L.~Qiao, W.~Chen, L.~Zhou, and J.~Zhang.
\newblock Learning deep correspondence through prior and posterior feature
  constancy.
\newblock In {\em arXiv:1712.01039}, 2017.

\bibitem{liu2016pami:depth}
F.~Liu, C.~Shen, G.~Lin, and I.~Reid.
\newblock Learning depth from single monocular images using deep convolutional
  neural fields.
\newblock {\em PAMI}, 38(10):2024--2039, Oct. 2016.

\bibitem{mayer2016cvpr:dispnet}
N.~Mayer, E.~Ilg, P.~H{\"a}usser, P.~Fischer, D.~Cremers, A.~Dosovitskiy, and
  T.~Brox.
\newblock A large dataset to train convolutional networks for disparity,
  optical flow, and scene flow estimation.
\newblock In {\em CVPR}, 2016.

\bibitem{palmisano2010jov}
S.~Palmisano, B.~Gillam, D.~G. Govan, R.~S. Allison, and J.~M. Harris.
\newblock Stereoscopic perception of real depths at large distances.
\newblock {\em Journal of Vision}, 10(6), June 2010.

\bibitem{pang2017arx:crl}
J.~Pang, W.~Sun, J.~Ren, C.~Yang, and Y.~Qiong.
\newblock Cascade residual learning: {A} two-stage convolutional neural network
  for stereo matching.
\newblock In {\em arXiv:1708.09204}, 2017.

\bibitem{seki2016bmvc:patch}
A.~Seki and M.~Pollefeys.
\newblock Patch based confidence prediction for dense disparity map.
\newblock In {\em British Machine Vision Conference (BMVC)}, 2016.

\bibitem{seki2017cvpr:sgmnet}
A.~Seki and M.~Pollefeys.
\newblock {SGM-Nets}: {S}emi-global matching with neural networks.
\newblock In {\em CVPR}, 2017.

\bibitem{shaked2017cvpr:highway}
A.~Shaked and L.~Wolf.
\newblock Improved stereo matching with constant highway networks and
  reflective loss.
\newblock In {\em CVPR}, 2017.

\bibitem{zbontar2016jmlr:stereo}
J.~\v{Z}bontar and Y.~LeCun.
\newblock Stereo matching by training a convolutional neural network to compare
  image patches.
\newblock {\em JMLR}, 17(65):1--32, 2016.

\bibitem{zhao2017itci:ssim}
H.~Zhao, O.~Gallo, I.~Frosio, and J.~Kautz.
\newblock Loss functions for image restoration with neural networks.
\newblock {\em IEEE Transactions on Computational Imaging}, 3(1), Mar. 2017.

\bibitem{zhong2017arx:ssl}
Y.~Zhong, Y.~Dai, and H.~Li.
\newblock Self-supervised learning for stereo matching with self-improving
  ability.
\newblock In {\em arXiv:1709.00930}, 2017.

\bibitem{zhou2004itip:ssim}
W.~Zhou, A.~C. Bovik, H.~R. Sheikh, and E.~P. Simoncelli.
\newblock Image quality assessment: {F}rom error visibility to structural
  similarity.
\newblock {\em IEEE Transactions on Image Processing}, 13(4):600--612, Apr.
  2004.

\end{thebibliography}
}

\vfill

\pagebreak

\appendix

\section{Network Architecture}

Tables~\ref{tab:netarch_baseline}--\ref{tab:netarch_tiny} provide the details of the network architectures used in the experiments of this paper.  The first table shows our baseline architecture, whereas the others show variations of the baseline (with \tbred{red} indicating the differences).  Note that these tables only describe the architecture for one of the two towers (left / right).  This is sufficient for inference, since only one tower is used for all architectures.  However, most implementations (that is, all except the single tower variant) contain two instances of the network for training.  More specifically, during training, all networks contain left and right instances of layers 1--10; the single tower variant contains only a single instance of the remaining layers, whereas all other variants contain two instances of these remaining layers.  (As described in the paper, $C=3$ is the number of color channels, and $F=32$ is the number of features.)

\vfill
\pagebreak

\begin{table}[h]
\begin{scriptsize}
\begin{center}
\caption{Our baseline network architecture.}
\label{tab:netarch_baseline}
\begin{tabular}{l|l|l}
	& \textbf{Layer description} &	\textbf{Output dimensions} \\
 & \emph{Input image (left or right)}	& $H \times W \times C$  \\
\hline 
	\hline
	\multicolumn{3}{c}{\textbf{2D Feature extraction:}}	 \\
1	& 2D conv, $5 \stimes 5$, stride 2, 32 features, ELU	& \sfrac{1}{2}$(H \times W) \times F$ \\
	\hline
2a 	& 2D conv, $3 \stimes 3$, stride 1, 32 features, ELU	& \sfrac{1}{2}$(H \times W) \times F$ \\
2b	& 2D conv, $3 \stimes 3$, stride 1, 32 features (no ELU) & \sfrac{1}{2}$(H \times W) \times F$ \\
   	& Add input of 2a and output of 2b, ELU	& \sfrac{1}{2}$(H \times W) \times F$ \\
	\hline
3a-9c	& Repeat 7 times:  2a, 2b, and addition & \sfrac{1}{2}$(H \times W) \times F$ \\
\hline 
10	& 2D conv, $3 \stimes 3$, stride 1, 32 features (no ELU) & \sfrac{1}{2}$(H \times W) \times F$ \\
\hline 
	\hline
	\multicolumn{3}{c}{\textbf{Cost volume:}}	 \\
11	& Concatenate feature maps from both towers & \sfrac{1}{2}$(D \stimes H \stimes W) \stimes 2F$ \\
\hline 
	\hline
	\multicolumn{3}{c}{\textbf{Stereo matching:}}	 \\
12a	& 3D conv, $3 \stimes 3 \stimes 3$, stride 1, 32 features, ELU	& \sfrac{1}{2}$(D \stimes H \stimes W) \stimes F$ \\
12b	& 3D conv, $3 \stimes 3 \stimes 3$, stride 1, 32 features, ELU	& \sfrac{1}{2}$(D \stimes H \stimes W) \stimes F$ \\
12c	& 3D conv, $3 \stimes 3 \stimes 3$, stride 2, 64 features, ELU	& \sfrac{1}{4}$(D \stimes H \stimes W) \stimes 2F$ \\
\hline 
13a	& 3D conv, $3 \stimes 3 \stimes 3$, stride 1, 64 features, ELU	& \sfrac{1}{4}$(D \stimes H \stimes W) \stimes 2F$ \\
13b	& 3D conv, $3 \stimes 3 \stimes 3$, stride 1, 64 features, ELU	& \sfrac{1}{4}$(D \stimes H \stimes W) \stimes 2F$ \\
13c	& 3D conv, $3 \stimes 3 \stimes 3$, stride 2, 64 features, ELU	& \sfrac{1}{8}$(D \stimes H \stimes W) \stimes 2F$ \\
\hline 
14a	& 3D conv, $3 \stimes 3 \stimes 3$, stride 1, 64 features, ELU	& \sfrac{1}{8}$(D \stimes H \stimes W) \stimes 2F$ \\
14b	& 3D conv, $3 \stimes 3 \stimes 3$, stride 1, 64 features, ELU	& \sfrac{1}{8}$(D \stimes H \stimes W) \stimes 2F$ \\
14c	& 3D conv, $3 \stimes 3 \stimes 3$, stride 2, 64 features, ELU	& \sfrac{1}{16}$(D \stimes H \stimes W) \stimes 2F$ \\
\hline 
15a	& 3D conv, $3 \stimes 3 \stimes 3$, stride 1, 64 features, ELU	& \sfrac{1}{16}$(D \stimes H \stimes W) \stimes 2F$ \\
15b	& 3D conv, $3 \stimes 3 \stimes 3$, stride 1, 64 features, ELU	& \sfrac{1}{16}$(D \stimes H \stimes W) \stimes 2F$ \\
15c	& 3D conv, $3 \stimes 3 \stimes 3$, stride 2, 128 features, ELU	& \sfrac{1}{32}$(D \stimes H \stimes W) \stimes 4F$ \\
\hline 
16	& 3D conv, $3 \stimes 3 \stimes 3$, stride 1, 128 features, ELU	& \sfrac{1}{32}$(D \stimes H \stimes W) \stimes 4F$ \\
17	& 3D conv, $3 \stimes 3 \stimes 3$, stride 1, 128 features, ELU	& \sfrac{1}{32}$(D \stimes H \stimes W) \stimes 4F$ \\
\hline 
18	& 3D deconv, $3 \stimes 3 \stimes 3$, stride 2, 64 features, ELU & \sfrac{1}{16}$(D \stimes H \stimes W) \stimes 2F$ \\
	& Add output of 15b and output of 18, ELU & \sfrac{1}{16}$(D \stimes H \stimes W) \stimes 2F$  \\
19	& 3D deconv, $3 \stimes 3 \stimes 3$, stride 2, 64 features, ELU	& \sfrac{1}{8}$(D \stimes H \stimes W) \stimes 2F$  \\
	& Add output of 14b and output of 19, ELU	& \sfrac{1}{8}$(D \stimes H \stimes W) \stimes 2F$ \\
20	& 3D deconv, $3 \stimes 3 \stimes 3$, stride 2, 64 features, ELU	& \sfrac{1}{4}$(D \stimes H \stimes W) \stimes 2F$  \\
	& Add output of 13b and output of 20, ELU	& \sfrac{1}{4}$(D \stimes H \stimes W) \stimes 2F$ \\
21	& 3D deconv, $3 \stimes 3 \stimes 3$, stride 2, 32 features, ELU	& \sfrac{1}{2}$(D \stimes H \stimes W) \stimes F$  \\
	& Add output of 12b and output of 21, ELU	& \sfrac{1}{2}$(D \stimes H \stimes W) \stimes F$ \\
\hline 
\hline 
	\multicolumn{3}{c}{\textbf{Upsampler:}}	 \\
22	& 3D deconv, $3 \stimes 3$, stride 2, 1 feature (no ELU) & $D \times H \times W \times 1$ \\
	\hline
\hline 
	\multicolumn{3}{c}{\textbf{Aggregator (Soft argmax):}}	 \\
23  & Reshape & $H \times W \times D$ \\
24  & Softargmax & $H \times W \times 1$
\end{tabular}
\end{center}
\end{scriptsize}
\end{table}

\vfill
\pagebreak

\begin{table}[h]
\begin{scriptsize}
\begin{center}
\caption{Our ML-argmax network architecture.}
\label{tab:netarch_mlargmax}
\begin{tabular}{l|l|l}
	& \textbf{Layer description} &	\textbf{Output dimensions} \\
 & \emph{Input image (left or right)}	& $H \times W \times C$  \\
\hline 
	\hline
	\multicolumn{3}{c}{\textbf{2D Feature extraction:}}	 \\
1	& 2D conv, $5 \stimes 5$, stride 2, 32 features, ELU	& \sfrac{1}{2}$(H \times W) \times F$ \\
	\hline
2a 	& 2D conv, $3 \stimes 3$, stride 1, 32 features, ELU	& \sfrac{1}{2}$(H \times W) \times F$ \\
2b	& 2D conv, $3 \stimes 3$, stride 1, 32 features (no ELU) & \sfrac{1}{2}$(H \times W) \times F$ \\
   	& Add input of 2a and output of 2b, ELU	& \sfrac{1}{2}$(H \times W) \times F$ \\
	\hline
3a-9c	& Repeat 7 times:  2a, 2b, and addition & \sfrac{1}{2}$(H \times W) \times F$ \\
\hline 
10	& 2D conv, $3 \stimes 3$, stride 1, 32 features (no ELU) & \sfrac{1}{2}$(H \times W) \times F$ \\
\hline 
	\hline
	\multicolumn{3}{c}{\textbf{Cost volume:}}	 \\
11	& Concatenate feature maps from both towers & \sfrac{1}{2}$(D \stimes H \stimes W) \stimes 2F$ \\
\hline 
	\hline
	\multicolumn{3}{c}{\textbf{Stereo matching:}}	 \\
12a	& 3D conv, $3 \stimes 3 \stimes 3$, stride 1, 32 features, ELU	& \sfrac{1}{2}$(D \stimes H \stimes W) \stimes F$ \\
12b	& 3D conv, $3 \stimes 3 \stimes 3$, stride 1, 32 features, ELU	& \sfrac{1}{2}$(D \stimes H \stimes W) \stimes F$ \\
12c	& 3D conv, $3 \stimes 3 \stimes 3$, stride 2, 64 features, ELU	& \sfrac{1}{4}$(D \stimes H \stimes W) \stimes 2F$ \\
\hline 
13a	& 3D conv, $3 \stimes 3 \stimes 3$, stride 1, 64 features, ELU	& \sfrac{1}{4}$(D \stimes H \stimes W) \stimes 2F$ \\
13b	& 3D conv, $3 \stimes 3 \stimes 3$, stride 1, 64 features, ELU	& \sfrac{1}{4}$(D \stimes H \stimes W) \stimes 2F$ \\
13c	& 3D conv, $3 \stimes 3 \stimes 3$, stride 2, 64 features, ELU	& \sfrac{1}{8}$(D \stimes H \stimes W) \stimes 2F$ \\
\hline 
14a	& 3D conv, $3 \stimes 3 \stimes 3$, stride 1, 64 features, ELU	& \sfrac{1}{8}$(D \stimes H \stimes W) \stimes 2F$ \\
14b	& 3D conv, $3 \stimes 3 \stimes 3$, stride 1, 64 features, ELU	& \sfrac{1}{8}$(D \stimes H \stimes W) \stimes 2F$ \\
14c	& 3D conv, $3 \stimes 3 \stimes 3$, stride 2, 64 features, ELU	& \sfrac{1}{16}$(D \stimes H \stimes W) \stimes 2F$ \\
\hline 
15a	& 3D conv, $3 \stimes 3 \stimes 3$, stride 1, 64 features, ELU	& \sfrac{1}{16}$(D \stimes H \stimes W) \stimes 2F$ \\
15b	& 3D conv, $3 \stimes 3 \stimes 3$, stride 1, 64 features, ELU	& \sfrac{1}{16}$(D \stimes H \stimes W) \stimes 2F$ \\
15c	& 3D conv, $3 \stimes 3 \stimes 3$, stride 2, 128 features, ELU	& \sfrac{1}{32}$(D \stimes H \stimes W) \stimes 4F$ \\
\hline 
16	& 3D conv, $3 \stimes 3 \stimes 3$, stride 1, 128 features, ELU	& \sfrac{1}{32}$(D \stimes H \stimes W) \stimes 4F$ \\
17	& 3D conv, $3 \stimes 3 \stimes 3$, stride 1, 128 features, ELU	& \sfrac{1}{32}$(D \stimes H \stimes W) \stimes 4F$ \\
\hline 
18	& 3D deconv, $3 \stimes 3 \stimes 3$, stride 2, 64 features, ELU & \sfrac{1}{16}$(D \stimes H \stimes W) \stimes 2F$ \\
	& Add output of 15b and output of 18, ELU & \sfrac{1}{16}$(D \stimes H \stimes W) \stimes 2F$  \\
19	& 3D deconv, $3 \stimes 3 \stimes 3$, stride 2, 64 features, ELU	& \sfrac{1}{8}$(D \stimes H \stimes W) \stimes 2F$  \\
	& Add output of 14b and output of 19, ELU	& \sfrac{1}{8}$(D \stimes H \stimes W) \stimes 2F$ \\
20	& 3D deconv, $3 \stimes 3 \stimes 3$, stride 2, 64 features, ELU	& \sfrac{1}{4}$(D \stimes H \stimes W) \stimes 2F$  \\
	& Add output of 13b and output of 20, ELU	& \sfrac{1}{4}$(D \stimes H \stimes W) \stimes 2F$ \\
21	& 3D deconv, $3 \stimes 3 \stimes 3$, stride 2, 32 features, ELU	& \sfrac{1}{2}$(D \stimes H \stimes W) \stimes F$  \\
	& Add output of 12b and output of 21, ELU	& \sfrac{1}{2}$(D \stimes H \stimes W) \stimes F$ \\
\hline 
\hline 
	\multicolumn{3}{c}{\textbf{Upsampler:}}	 \\
22	& 3D deconv, $3 \stimes 3$, stride 2, 1 feature (no ELU) & $D \times H \times W \times 1$ \\
	\hline
\hline 
	\multicolumn{3}{c}{\textbf{Aggregator (Machine-learned argmax):}}	 \\
23  & Reshape & $H \times W \times D$ \\
\tbred{24}	& \tbred{2D conv, $3 \stimes 3$, stride 1, D feature, ELU}	& \tbred{$H \times W \times D$} \\
\tbred{25}	& \tbred{2D conv, $3 \stimes 3$, stride 1, D feature, ELU}	& \tbred{$H \times W \times D$} \\
\tbred{26}	& \tbred{2D conv, $3 \stimes 3$, stride 1, D feature, ELU}	& \tbred{$H \times W \times D$} \\
\tbred{27}	& \tbred{2D conv, $3 \stimes 3$, stride 1, D feature, ELU}	& \tbred{$H \times W \times D$} \\
\tbred{28}	& \tbred{2D conv, $3 \stimes 3$, stride 1, 1 feature, sigmoid}	& \tbred{$H \times W \times 1$}
\end{tabular}
\end{center}
\end{scriptsize}
\end{table}

\vfill
\pagebreak

\begin{table}[h]
\begin{scriptsize}
\begin{center}
\caption{Correlation network architecture.}
\label{tab:netarch_corr}
\begin{tabular}{l|l|l}
	& \textbf{Layer description} &	\textbf{Output dimensions} \\
 & \emph{Input image (left or right)}	& $H \times W \times C$  \\
\hline 
	\hline
	\multicolumn{3}{c}{\textbf{2D Feature extraction:}}	 \\
1	& 2D conv, $5 \stimes 5$, stride 2, 32 features, ELU	& \sfrac{1}{2}$(H \times W) \times F$ \\
	\hline
2a 	& 2D conv, $3 \stimes 3$, stride 1, 32 features, ELU	& \sfrac{1}{2}$(H \times W) \times F$ \\
2b	& 2D conv, $3 \stimes 3$, stride 1, 32 features (no ELU) & \sfrac{1}{2}$(H \times W) \times F$ \\
   	& Add input of 2a and output of 2b, ELU	& \sfrac{1}{2}$(H \times W) \times F$ \\
	\hline
3a-9c	& Repeat 7 times:  2a, 2b, and addition & \sfrac{1}{2}$(H \times W) \times F$ \\
\hline 
10	& 2D conv, $3 \stimes 3$, stride 1, 32 features (no ELU) & \sfrac{1}{2}$(H \times W) \times F$ \\
\hline 
	\hline
	\multicolumn{3}{c}{\textbf{Cost volume:}}	 \\
\tbred{11}	& \tbred{Correlate feature maps from both towers} & \tbred{\sfrac{1}{2}$(D \stimes H \stimes W) \stimes 1$} \\
\hline 
	\hline
	\multicolumn{3}{c}{\textbf{Stereo matching:}}	 \\
12a	& 3D conv, $3 \stimes 3 \stimes 3$, stride 1, 32 features, ELU	& \sfrac{1}{2}$(D \stimes H \stimes W) \stimes F$ \\
12b	& 3D conv, $3 \stimes 3 \stimes 3$, stride 1, 32 features, ELU	& \sfrac{1}{2}$(D \stimes H \stimes W) \stimes F$ \\
12c	& 3D conv, $3 \stimes 3 \stimes 3$, stride 2, 64 features, ELU	& \sfrac{1}{4}$(D \stimes H \stimes W) \stimes 2F$ \\
\hline 
13a	& 3D conv, $3 \stimes 3 \stimes 3$, stride 1, 64 features, ELU	& \sfrac{1}{4}$(D \stimes H \stimes W) \stimes 2F$ \\
13b	& 3D conv, $3 \stimes 3 \stimes 3$, stride 1, 64 features, ELU	& \sfrac{1}{4}$(D \stimes H \stimes W) \stimes 2F$ \\
13c	& 3D conv, $3 \stimes 3 \stimes 3$, stride 2, 64 features, ELU	& \sfrac{1}{8}$(D \stimes H \stimes W) \stimes 2F$ \\
\hline 
14a	& 3D conv, $3 \stimes 3 \stimes 3$, stride 1, 64 features, ELU	& \sfrac{1}{8}$(D \stimes H \stimes W) \stimes 2F$ \\
14b	& 3D conv, $3 \stimes 3 \stimes 3$, stride 1, 64 features, ELU	& \sfrac{1}{8}$(D \stimes H \stimes W) \stimes 2F$ \\
14c	& 3D conv, $3 \stimes 3 \stimes 3$, stride 2, 64 features, ELU	& \sfrac{1}{16}$(D \stimes H \stimes W) \stimes 2F$ \\
\hline 
15a	& 3D conv, $3 \stimes 3 \stimes 3$, stride 1, 64 features, ELU	& \sfrac{1}{16}$(D \stimes H \stimes W) \stimes 2F$ \\
15b	& 3D conv, $3 \stimes 3 \stimes 3$, stride 1, 64 features, ELU	& \sfrac{1}{16}$(D \stimes H \stimes W) \stimes 2F$ \\
15c	& 3D conv, $3 \stimes 3 \stimes 3$, stride 2, 128 features, ELU	& \sfrac{1}{32}$(D \stimes H \stimes W) \stimes 4F$ \\
\hline 
16	& 3D conv, $3 \stimes 3 \stimes 3$, stride 1, 128 features, ELU	& \sfrac{1}{32}$(D \stimes H \stimes W) \stimes 4F$ \\
17	& 3D conv, $3 \stimes 3 \stimes 3$, stride 1, 128 features, ELU	& \sfrac{1}{32}$(D \stimes H \stimes W) \stimes 4F$ \\
\hline 
18	& 3D deconv, $3 \stimes 3 \stimes 3$, stride 2, 64 features, ELU & \sfrac{1}{16}$(D \stimes H \stimes W) \stimes 2F$ \\
	& Add output of 15b and output of 18, ELU & \sfrac{1}{16}$(D \stimes H \stimes W) \stimes 2F$  \\
19	& 3D deconv, $3 \stimes 3 \stimes 3$, stride 2, 64 features, ELU	& \sfrac{1}{8}$(D \stimes H \stimes W) \stimes 2F$  \\
	& Add output of 14b and output of 19, ELU	& \sfrac{1}{8}$(D \stimes H \stimes W) \stimes 2F$ \\
20	& 3D deconv, $3 \stimes 3 \stimes 3$, stride 2, 64 features, ELU	& \sfrac{1}{4}$(D \stimes H \stimes W) \stimes 2F$  \\
	& Add output of 13b and output of 20, ELU	& \sfrac{1}{4}$(D \stimes H \stimes W) \stimes 2F$ \\
21	& 3D deconv, $3 \stimes 3 \stimes 3$, stride 2, 32 features, ELU	& \sfrac{1}{2}$(D \stimes H \stimes W) \stimes F$  \\
	& Add output of 12b and output of 21, ELU	& \sfrac{1}{2}$(D \stimes H \stimes W) \stimes F$ \\
\hline 
\hline 
	\multicolumn{3}{c}{\textbf{Upsampler:}}	 \\
22	& 3D deconv, $3 \stimes 3$, stride 2, 1 feature (no ELU) & $D \times H \times W \times 1$ \\
	\hline
\hline 
	\multicolumn{3}{c}{\textbf{Aggregator (Soft argmax):}}	 \\
23  & Reshape & $H \times W \times D$ \\
24  & Softargmax & $H \times W \times 1$
\end{tabular}
\end{center}
\end{scriptsize}
\end{table}

\vfill
\pagebreak

\begin{table}[h]
\begin{scriptsize}
\begin{center}
\caption{No bottleneck network architecture.}
\label{tab:netarch_nobottle}
\begin{tabular}{l|l|l}
\\
	& \textbf{Layer description} &	\textbf{Output dimensions} \\
 & \emph{Input image (left or right)}	& $H \times W \times C$  \\
\hline 
	\hline
	\multicolumn{3}{c}{\textbf{2D Feature extraction:}}	 \\
1	& 2D conv, $5 \stimes 5$, stride 2, 32 features, ELU	& \sfrac{1}{2}$(H \times W) \times F$ \\
	\hline
2a 	& 2D conv, $3 \stimes 3$, stride 1, 32 features, ELU	& \sfrac{1}{2}$(H \times W) \times F$ \\
2b	& 2D conv, $3 \stimes 3$, stride 1, 32 features (no ELU) & \sfrac{1}{2}$(H \times W) \times F$ \\
   	& Add input of 2a and output of 2b, ELU	& \sfrac{1}{2}$(H \times W) \times F$ \\
	\hline
3a-9c	& Repeat 7 times:  2a, 2b, and addition & \sfrac{1}{2}$(H \times W) \times F$ \\
\hline 
10	& 2D conv, $3 \stimes 3$, stride 1, 32 features (no ELU) & \sfrac{1}{2}$(H \times W) \times F$ \\
\hline 
	\hline
	\multicolumn{3}{c}{\textbf{Cost volume:}}	 \\
11	& Concatenate feature maps from both towers & \sfrac{1}{2}$(D \stimes H \stimes W) \stimes 2F$ \\
\hline 
	\hline
	\multicolumn{3}{c}{\textbf{Stereo matching:}}	 \\
\tbred{12a}	& & \\
\tbred{12b}	& \hspace{7em} \tbred{------------}	& \\
\tbred{12c}	& & \\
\hline 
\tbred{13a}	& & \\
\tbred{13b}	& \hspace{7em} \tbred{------------}	& \\
\tbred{13c}	& & \\
\hline 
\tbred{14a}	& & \\
\tbred{14b}	& \hspace{7em} \tbred{------------}	& \\
\tbred{14c}	& & \\
\hline 
\tbred{15a}	& & \\
\tbred{15b}	& \hspace{7em} \tbred{------------}	& \\
\tbred{15c}	& & \\
\hline 
16	& 3D conv, $3 \stimes 3 \stimes 3$, stride 1, 32 features, ELU	& \sfrac{1}{2}$(D \stimes H \stimes W) \stimes F$ \\
17	& 3D conv, $3 \stimes 3 \stimes 3$, stride 1, 32 features, ELU	& \sfrac{1}{2}$(D \stimes H \stimes W) \stimes F$ \\
\hline 
\tbred{18}	& \hspace{7em} \tbred{------------}	& \\
	& & \\
\tbred{19}	& \hspace{7em} \tbred{------------}	& \\
	& & \\
\tbred{20}	& \hspace{7em} \tbred{------------}	& \\
	& & \\
\tbred{21}	& \hspace{7em} \tbred{------------}	& \\
	& & \\
\hline 
\hline 
	\multicolumn{3}{c}{\textbf{Upsampler:}}	 \\
22	& 3D deconv, $3 \stimes 3$, stride 2, 1 feature (no ELU) & $D \times H \times W \times 1$ \\
	\hline
\hline 
	\multicolumn{3}{c}{\textbf{Aggregator (Soft argmax):}}	 \\
23  & Reshape & $H \times W \times D$ \\
24  & Softargmax & $H \times W \times 1$
\end{tabular}
\end{center}
\end{scriptsize}
\end{table}

\vfill
\pagebreak

\begin{table}[h]
\begin{scriptsize}
\begin{center}
\caption{Small network architecture.}
\label{tab:netarch_small}
\begin{tabular}{l|l|l}
\\
	& \textbf{Layer description} &	\textbf{Output dimensions} \\
 & \emph{Input image (left or right)}	& $H \times W \times C$  \\
\hline 
	\hline
	\multicolumn{3}{c}{\textbf{2D Feature extraction:}}	 \\
1	& 2D conv, $5 \stimes 5$, stride 2, 32 features, ELU	& \sfrac{1}{2}$(H \times W) \times F$ \\
	\hline
2 	& 2D conv, $3 \stimes 3$, stride 1, 32 features, ELU	& \sfrac{1}{2}$(H \times W) \times F$ \\
3 	& 2D conv, $3 \stimes 3$, stride 1, 32 features, ELU	& \sfrac{1}{2}$(H \times W) \times F$ \\
4 	& 2D conv, $3 \stimes 3$, stride 1, 32 features, ELU	& \sfrac{1}{2}$(H \times W) \times F$ \\
5 	& 2D conv, $3 \stimes 3$, stride 1, 32 features, ELU	& \sfrac{1}{2}$(H \times W) \times F$ \\
\\
\hline 
	\hline
	\multicolumn{3}{c}{\textbf{Cost volume:}}	 \\
11	& Concatenate feature maps from both towers & \sfrac{1}{2}$(D \stimes H \stimes W) \stimes 2F$ \\
\hline 
	\hline
	\multicolumn{3}{c}{\textbf{Stereo matching:}}	 \\
12a	& 3D conv, $3 \stimes 3 \stimes 3$, stride 1, 32 features, ELU	& \sfrac{1}{2}$(D \stimes H \stimes W) \stimes F$ \\
12b	& 3D conv, $3 \stimes 3 \stimes 3$, stride 1, 32 features, ELU	& \sfrac{1}{2}$(D \stimes H \stimes W) \stimes F$ \\
12c	& 3D conv, $3 \stimes 3 \stimes 3$, stride 2, 64 features, ELU	& \sfrac{1}{4}$(D \stimes H \stimes W) \stimes 2F$ \\
\hline 
13a	& 3D conv, $3 \stimes 3 \stimes 3$, stride 1, 64 features, ELU	& \sfrac{1}{4}$(D \stimes H \stimes W) \stimes 2F$ \\
13b	& 3D conv, $3 \stimes 3 \stimes 3$, stride 1, 64 features, ELU	& \sfrac{1}{4}$(D \stimes H \stimes W) \stimes 2F$ \\
13c	& 3D conv, $3 \stimes 3 \stimes 3$, stride 2, 128 features, ELU	& \sfrac{1}{8}$(D \stimes H \stimes W) \stimes 4F$ \\
\hline 
\tbred{14a}	& & \\
\tbred{14b}	& \hspace{7em} \tbred{------------}	& \\
\tbred{14c}	& & \\
\hline 
\tbred{15a}	& & \\
\tbred{15b}	& \hspace{7em} \tbred{------------}	& \\
\tbred{15c}	& & \\
\hline 
16	& 3D conv, $3 \stimes 3 \stimes 3$, stride 1, 128 features, ELU	& \sfrac{1}{8}$(D \stimes H \stimes W) \stimes 4F$ \\
17	& 3D conv, $3 \stimes 3 \stimes 3$, stride 1, 128 features, ELU	& \sfrac{1}{8}$(D \stimes H \stimes W) \stimes 4F$ \\
\hline 
18	& 3D deconv, $3 \stimes 3 \stimes 3$, stride 2, 64 features, ELU & \sfrac{1}{4}$(D \stimes H \stimes W) \stimes 2F$ \\
	& Add output of 13b and output of 18, ELU & \sfrac{1}{4}$(D \stimes H \stimes W) \stimes 2F$  \\
19	& 3D deconv, $3 \stimes 3 \stimes 3$, stride 2, 32 features, ELU	& \sfrac{1}{2}$(D \stimes H \stimes W) \stimes F$  \\
	& Add output of 12b and output of 19, ELU	& \sfrac{1}{2}$(D \stimes H \stimes W) \stimes F$ \\
\tbred{20}	& \hspace{7em} \tbred{------------}	& \\
	& & \\
\tbred{21}	& \hspace{7em} \tbred{------------}	& \\
	& & \\
\hline 
\hline 
	\multicolumn{3}{c}{\textbf{Upsampler:}}	 \\
22	& 3D deconv, $3 \stimes 3$, stride 2, 1 feature (no ELU) & $D \times H \times W \times 1$ \\
	\hline
\hline 
	\multicolumn{3}{c}{\textbf{Aggregator (Soft argmax):}}	 \\
23  & Reshape & $H \times W \times D$ \\
24  & Softargmax & $H \times W \times 1$
\end{tabular}
\end{center}
\end{scriptsize}
\end{table}

\vfill
\pagebreak

\begin{table}[h]
\begin{scriptsize}
\begin{center}
\caption{Tiny network architecture.}
\label{tab:netarch_tiny}
\begin{tabular}{l|l|l}
	& \textbf{Layer description} &	\textbf{Output dimensions} \\
 & \emph{Input image (left or right)}	& $H \times W \times C$  \\
\hline 
	\hline
	\multicolumn{3}{c}{\textbf{2D Feature extraction:}}	 \\
1	& 2D conv, $5 \stimes 5$, stride 2, 32 features, ELU	& \sfrac{1}{2}$(H \times W) \times F$ \\
	\hline
2 	& 2D conv, $3 \stimes 3$, stride 1, 32 features, ELU	& \sfrac{1}{2}$(H \times W) \times F$ \\
3 	& 2D conv, $3 \stimes 3$, stride 1, 32 features, ELU	& \sfrac{1}{2}$(H \times W) \times F$ \\
4 	& 2D conv, $3 \stimes 3$, stride 1, 32 features, ELU	& \sfrac{1}{2}$(H \times W) \times F$ \\
5 	& 2D conv, $3 \stimes 3$, stride 1, 32 features, ELU	& \sfrac{1}{2}$(H \times W) \times F$ \\
\\
\hline 
	\hline
	\multicolumn{3}{c}{\textbf{Cost volume:}}	 \\
11	& Concatenate feature maps from both towers & \sfrac{1}{2}$(D \stimes H \stimes W) \stimes 2F$ \\
\hline 
	\hline
	\multicolumn{3}{c}{\textbf{Stereo matching:}}	 \\
12a	& 3D conv, $3 \stimes 3 \stimes 3$, stride 1, 16 features, ELU	& \sfrac{1}{2}$(D \stimes H \stimes W) \stimes \Fhalf$ \\
12b	& 3D conv, $3 \stimes 3 \stimes 3$, stride 1, 16 features, ELU	& \sfrac{1}{2}$(D \stimes H \stimes W) \stimes \Fhalf$ \\
12c	& 3D conv, $3 \stimes 3 \stimes 3$, stride 2, 32 features, ELU	& \sfrac{1}{4}$(D \stimes H \stimes W) \stimes F$ \\
\hline 
13a	& 3D conv, $3 \stimes 3 \stimes 3$, stride 1, 32 features, ELU	& \sfrac{1}{4}$(D \stimes H \stimes W) \stimes F$ \\
13b	& 3D conv, $3 \stimes 3 \stimes 3$, stride 1, 32 features, ELU	& \sfrac{1}{4}$(D \stimes H \stimes W) \stimes F$ \\
13c	& 3D conv, $3 \stimes 3 \stimes 3$, stride 2, 64 features, ELU	& \sfrac{1}{8}$(D \stimes H \stimes W) \stimes 2F$ \\
\hline 
\tbred{14a}	& & \\
\tbred{14b}	& \hspace{7em} \tbred{------------}	& \\
\tbred{14c}	& & \\
\hline 
\tbred{15a}	& & \\
\tbred{15b}	& \hspace{7em} \tbred{------------}	& \\
\tbred{15c}	& & \\
\hline 
16	& 3D conv, $3 \stimes 3 \stimes 3$, stride 1, 64 features, ELU	& \sfrac{1}{8}$(D \stimes H \stimes W) \stimes 2F$ \\
17	& 3D conv, $3 \stimes 3 \stimes 3$, stride 1, 64 features, ELU	& \sfrac{1}{8}$(D \stimes H \stimes W) \stimes 2F$ \\
\hline 
18	& 3D deconv, $3 \stimes 3 \stimes 3$, stride 2, 32 features, ELU & \sfrac{1}{4}$(D \stimes H \stimes W) \stimes F$ \\
	& Add output of 13b and output of 18, ELU & \sfrac{1}{4}$(D \stimes H \stimes W) \stimes F$  \\
19	& 3D deconv, $3 \stimes 3 \stimes 3$, stride 2, 16 features, ELU	& \sfrac{1}{2}$(D \stimes H \stimes W) \stimes \Fhalf$  \\
	& Add output of 12b and output of 19, ELU	& \sfrac{1}{2}$(D \stimes H \stimes W) \stimes \Fhalf$ \\
\tbred{20}	& \hspace{7em} \tbred{------------}	& \\
	& & \\
\tbred{21}	& \hspace{7em} \tbred{------------}	& \\
	& & \\
\hline 
\hline 
	\multicolumn{3}{c}{\textbf{Upsampler:}}	 \\
22	& 3D deconv, $3 \stimes 3$, stride 2, 1 feature (no ELU) & $D \times H \times W \times 1$ \\
	\hline
\hline 
	\multicolumn{3}{c}{\textbf{Aggregator (Soft argmax):}}	 \\
23  & Reshape & $H \times W \times D$ \\
24  & Softargmax & $H \times W \times 1$
\end{tabular}
\end{center}
\end{scriptsize}
\end{table}

\vfill
\pagebreak

\end{document}